\documentclass{article} % For LaTeX2e
\usepackage{iclr2015,times}
\usepackage{hyperref}
\usepackage{url}

\usepackage{amsmath, amsfonts, amssymb, amsthm, mathtools, calrsfs}
\usepackage[pdftex]{graphicx}
\graphicspath{{figures/}}

\usepackage{caption}
\usepackage{subcaption}

\usepackage{tikz}
\usetikzlibrary{bayesnet}

\usepackage{algorithm}
\usepackage{algorithmicx}
\usepackage[noend]{algpseudocode}

\DeclareMathOperator{\cord}{cord}

\newcommand{\bfv}{\ensuremath{\pmb{v}}}
\newcommand{\bfV}{\ensuremath{\pmb{V}}}
\newcommand{\bfh}{\ensuremath{\pmb{h}}}

\newcommand{\bfW}{\ensuremath{\pmb{W}}}
\newcommand{\bfS}{\ensuremath{\pmb{S}}}
\newcommand{\bfb}{\ensuremath{\pmb{b}}}
\newcommand{\bfc}{\ensuremath{\pmb{c}}}

\newcommand\indep{\protect\mathpalette{\protect\indeP}{\perp}}
\newcommand\nindep{\not\indep}
\def\indeP#1#2{\mathrel{\rlap{$#1#2$}\mkern2mu{#1#2}}}

\algnewcommand\algorithmicinput{\textbf{Input:}}
\algnewcommand\Input{\item[\algorithmicinput]}
\algnewcommand\algorithmicoutput{\textbf{Output:}}
\algnewcommand\Output{\item[\algorithmicoutput]}

\title{Learning Non-deterministic Representations with Energy-based Ensembles}

\author{
Maruan~Al-Shedivat\thanks{Corresponding author: \href{http://maruan.alshedivat.com}{maruan.alshedivat.com}.} \\
Computer, Electrical and Mathematical Sciences and Engineering Division\\
King Abdullah University of Science and Technology (KAUST)\\
Thuwal 23955-6900, Saudi Arabia \\
\texttt{maruan.shedivat@kaust.edu.sa} \\
\And
Emre Neftci and Gert Cauwenberghs \\
Institute for Neural Computation \\
University of California, San Diego \\
La Jolla, CA 92093, USA \\
\texttt{\{nemre,gert\}@ucsd.edu}\\
}

\iclrfinalcopy % Uncomment for camera-ready version

\begin{document}

\maketitle

\begin{abstract}
The goal of a generative model is to capture the distribution underlying the data, typically through latent variables.
After training, these variables are often used as a new representation, more effective than the original features in a variety of learning tasks.
However, the representations constructed by contemporary generative models are usually point-wise deterministic mappings from the original feature space.
Thus, even with representations robust to class-specific transformations, statistically driven models trained on them would not be able to generalize when the labeled data is scarce.
Inspired by the stochasticity of the synaptic connections in the brain, we introduce Energy-based Stochastic Ensembles.
These ensembles can learn non-deterministic representations, i.e., mappings from the feature space to a family of distributions in the latent space.
These mappings are encoded in a distribution over a (possibly infinite) collection of models.
By conditionally sampling models from the ensemble, we obtain multiple representations for every input example and effectively augment the data.
We propose an algorithm similar to contrastive divergence for training restricted Boltzmann stochastic ensembles.
Finally, we demonstrate the concept of the stochastic representations on a synthetic dataset as well as test them in the one-shot learning scenario on MNIST.

\end{abstract}

\section{Introduction}
Learning data representations is a powerful technique that has been widely adopted in many fields of artificial intelligence~\citep{bengio2009learning}.
Its main goal is usually to learn transformations of the data that disentangle different classes in the new space, that are robust to the noise in the input and tolerant to the variations along the class-specific manifolds \citep{dicarlo2007untangling}.
A widely used technique of constructing representations is based on using probabilistic generative models of the data.
Latent variables in such models can capture high-order correlations between the samples and are successful as new representations in a variety of tasks~\citep{bengio2009learning}.
However, even high quality representations do not solve the problem of generalization: statistically derived discriminative models require a sufficient number of labeled training examples in order to exhibit good performance when tested.

The standard way of ameliorating the problem of overfitting due to the limited training data is based on enforcing a regularization \citep{bishop2006pattern}.
\cite{maaten2013learning} recently demonstrated that the artificial data augmentation via feature corruption effectively plays the role of a data-adaptive regularization.
\cite{wager2013dropout} also showed that the dropout techniques applied to generalized linear models result into an adaptive regularization.
In other words, these approaches confirm that having more (even corrupted) training data is equivalent to regularizing the model.

On the other hand, instead of corrupting features one can try perform regularization by perturbing the learning model itself.
Parameter perturbation leads to the notion of (possibly infinite) collection of models that we call \textit{stochastic ensemble}.
The Dropout \citep{hinton2012dropout} and DropConnect \citep{wan2013regularization} techniques are successful examples of using a particular form of stochastic ensemble in the context of feedforward neural networks.
A unified framework for a collection of perturbed models (which also encompasses the data corruption methods) was recently introduced by \cite{bachman2014learning}:
An arbitrary noise process was used to perturb a parametric parent model to generate an ensemble of child models.
The training was performed by minimizing the marginalized loss function, and yielded an effectively regularized parent model that was robust to the introduced noise.

The above mentioned approaches use \textit{fixed} corruption processes to regularize the original model and improve its generalization.
Injection of arbitrary noise improves the robustness of the model to the corruption process \citep{maaten2013learning}, but it does not necessarily capture the information about the generative process behind the actual data.
In order to capture the variance of the data manifold, we propose to learn the noise that perturbs the model.
Our work is inspired by the adaptive stochasticity of the synaptic connections in the brain.
According to the experimental data, synapses between cortical neurons are highly unreliable \citep{branco2009probability}.
Moreover, this type of stochasticity is adaptive and adjusted by the learning \citep{stevens1994changes}.

In this paper, we introduce energy-based stochastic ensembles (EBSEs) which can be trained in unsupervised fashion to fit a data distribution.
These ensembles result from energy-based models (EBMs) with stochastic parameters.
The EBSE learning procedure optimizes the log-likelihood of the ensemble by tuning a parametric distribution over the models.
This distribution is first arbitrarily initialized, and then tuned by an expectation-maximization (EM) like procedure:
In the E-step, we perform sampling to estimate the necessary expectations, while in the M-step we maximize the log-likelihood with respect to the ensemble parameters.
We further develop an algorithm for learning restricted Boltzmann stochastic ensembles (RBSE) similar to contrastive divergence.

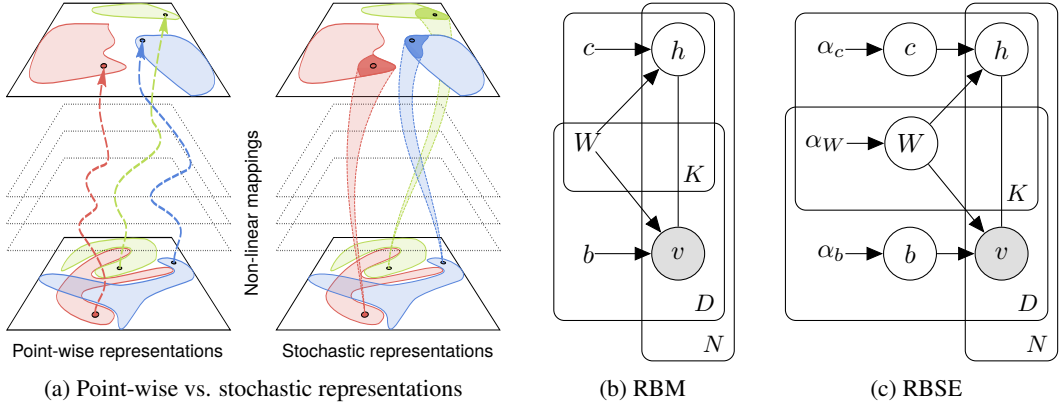
\begin{figure}[t!]
\centering
\begin{subfigure}[b]{0.47\textwidth}
\includegraphics[width=\textwidth]{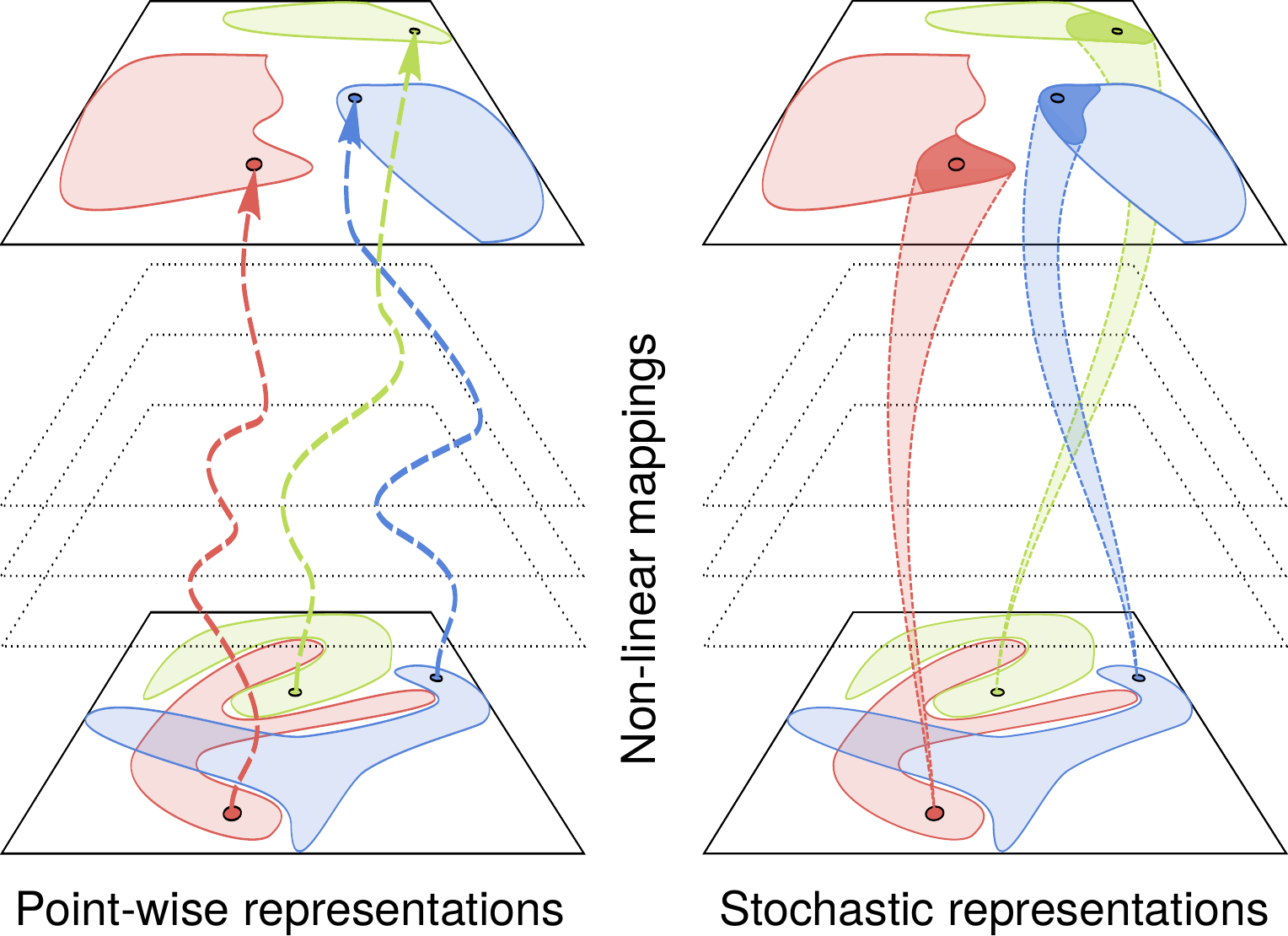}
\caption{Point-wise vs. stochastic representations}
\label{fig:point-wise_vs_stochastic}
\end{subfigure}
~~~~~~
\begin{subfigure}[b]{0.17\textwidth}
  \begin{tikzpicture}
    % Define nodes
    \node[obs]                                      (v) {$v$};
    \node[latent, above=2.0 of v]                   (h) {$h$};
    \node[const,  above=1.0 of v, xshift=-1.2cm]    (w) {$W$};
    \node[const,  left=0.75 of v]                   (b) {$b$};
    \node[const,  left=0.75 of h]                   (c) {$c$};

    % Connect the nodes
    \edge[-] {v}   {h} ; %
    \edge    {w,b} {v} ; %
    \edge    {w,c} {h} ; %

    % Plates
    \plate {hwc} {(h)(c)(w)}    {$K$} ;
    \plate {vwb} {(v)(b)(w)(hwc.south west)(hwc.south east)}    {$D$} ;
    \plate {}    {(v)(h)(hwc.north east)(vwb.south east)} {$N$} ;
  \end{tikzpicture}
  \caption{RBM}\label{fig:RBM}
\end{subfigure}
~~~~~~
\begin{subfigure}[b]{0.25\textwidth}
  \begin{tikzpicture}
    % Define nodes
    \node[obs]                                     (v) {$v$};
    \node[latent, above=0.75 of v, xshift=-1.2cm]  (w) {$W$};
    \node[latent, above=2.0 of v]                  (h) {$h$};
    \node[latent, left=0.5 of v]                   (b) {$b$};
    \node[latent, left=0.5 of h]                   (c) {$c$};
    \node[const,  left=0.5  of w]                  (aw) {$\alpha_W$};
    \node[const,  left=0.5  of b]                  (ab) {$\alpha_b$};
    \node[const,  left=0.5  of c]                  (ac) {$\alpha_c$};

    % Connect the nodes
    \edge[-] {v}   {h}      ; %
    \edge    {w}   {v,h}    ; %
    \edge    {b}   {v}      ; %
    \edge    {c}   {h}      ; %
    \edge    {aw}  {w}      ; %
    \edge    {ab}  {b}      ; %
    \edge    {ac}  {c}      ; %

    % Plates
    \plate {hwc} {(h)(w)(c)(aw)(ac)}                                  {$K$} ;
    \plate {vwb} {(v)(w)(b)(aw)(ab)(hwc.south west)(hwc.south east)}  {$D$} ;
    \plate {}  {(v)(h)(hwc.north east)(vwb.south east)}               {$N$} ;
  \end{tikzpicture}
  \caption{RBSE}\label{fig:RBSE}
\end{subfigure}
\caption{(a) The difference between point-wise and stochastic representations.
While both mappings disentangle the manifolds of different classes, stochasticity additionally captures the data manifold in the representation space.
(b) RBM and (c) RBSE graphical diagrams.
The shaded nodes are the visible variable, the black nodes are the latent variables, literals outside of nodes are constants (model parameters).
Plates denote variable scopes with sizes indicated at the bottom right corners.}
\end{figure}

Using a pre-trained ensemble, we further introduce the notion of non-deterministic representations: instead of constructing point-wise mappings from the original feature space to a new latent space, we propose using stochastic mappings (Figure~\ref{fig:point-wise_vs_stochastic}).
The stochasticity of the representations is based on the distribution stored in the ensemble.
For every input object we can sample multiple models, and hence perform non-deterministic transformations to obtain a set of different representations.
The ensemble is tuned to capture the variance of the data, hence the stochastic representations are likely to better capture the data manifold.
We demonstrate these concepts visually by performing experiments on a two-dimensional synthetic dataset.
Finally, we train a classifier on the representations of the MNIST hand-written digits generated by an RBSE and observe that the generalization capability in the one-shot learning scenario improves.

%% E: Are you planning to demonstrate in another way that the ``stochastic representations are likely to capture the data manifold''? One idea would be to estimate the data likelihood (= generative power) using annealed importance sampling.

\section{Energy-based Stochastic Ensembles}\label{sec:EBSE}
The distribution of the binary data vectors $\bfv \in \{0, 1\}^{D}$ can be encoded with the following energy-based model (EBM) that has some binary hidden variables $\bfh~\in~\{0, 1\}^{K}$:
\begin{equation}
\label{eq:EBM}
P(\bfv, \bfh; \theta) = \frac{e^{-E(\bfv, \bfh; \theta)}}{Z(\theta)},
\end{equation}
where $\theta$ denotes the model parameters, $E(\bfv, \bfh; \theta)$ is a parametric scalar energy function \citep{lecun2006tutorial}, and $Z(\theta)$ is the normalizing coefficient (partition function).
If we impose a distribution over the model parameters (i.e., perturb $\theta$ according to some noise generation process), we obtain an energy-based stochastic ensemble (EBSE).
In order to optimize the distribution over the models in the ensemble, we parametrize the noise generation process with $\alpha$:
\begin{equation}
\label{eq:EBSE}
P(\bfv, \bfh, \theta; \alpha) = P(\bfv, \bfh \mid \theta) P(\theta; \alpha) = \frac{e^{-E(\bfv, \bfh, \theta)}}{Z(\theta)} P(\theta; \alpha) = \frac{e^{-E(\bfv, \bfh, \theta) - \phi(\theta; \alpha)}}{\zeta(\alpha)},
\end{equation}
where $\phi(\theta; \alpha)$ is an additional energy potential, and $\zeta(\alpha)$ is a new partition function.
EBSE can be trained by maximizing the data log-likelihood $\log P(\bfV; \alpha)$ with respect to parameters $\alpha$, where $\bfV$ denotes the entire dataset.

The introduced form of the model \eqref{eq:EBSE} allows to think of $P(\theta; \alpha)$ as a prior.
Hence, we can relate the optimization of EBSE with the Bayesian inference framework for a standard EBM by taking expectation over the posterior $P(\theta \mid \bfV; \alpha)$:
\begin{equation}
\begin{aligned}
\label{eq:ll_EBSE}
\log P(\bfV; \alpha)
&= \int P(\theta \mid \bfV; \alpha) \log P(\bfV \mid \theta) d\theta - D_{KL}\left[P(\theta \mid \bfV; \alpha) \| P(\theta; \alpha) \right]\\
&= \underbrace{\mathbb{E} \left[ \log P(\bfV \mid \theta) \right]_{P(\theta \mid \bfV; \alpha)}}_\text{Expected EBM log-likelihood} - \underbrace{D_{KL}\left[P(\theta \mid \bfV; \alpha) \| P(\theta; \alpha) \right]}_\text{KL divergence of posterior and prior}.
\end{aligned}
\end{equation}
Since the KL divergence is non-negative, by optimizing the log-likelihood of EBSE, we effectively maximize a lower bound on the EBM log-likelihood averaged over the posterior $P(\theta \mid \bfV; \alpha)$.
This relates our approach to the feature corruption method \citep[Eq. 2]{maaten2013learning} which optimizes the expected loss directly, but over a simple posterior feature corruption distribution $P(\tilde\bfv \mid \bfv)$.
Eventually, once we trained the stochastic ensemble of energy-based generative models, we get new parameters $\hat\alpha$ that make $P(\theta; \hat\alpha)$ better tuned to capture the variance of the data.

Below, we derive the gradient of the EBSE log-likelihood \eqref{eq:ll_EBSE} for the general energy case.
Then, we focus on the restricted Boltzmann stochastic ensembles (RBSE), analyze their structure, and propose an efficient learning algorithm.

\subsection{Log-likelihood Optimization}\label{sec:ll_optimization}
The gradient of the log-likelihood function \eqref{eq:ll_EBSE} can be written in the following way:
\begin{equation}
\label{eq:grad_ll_EBSE}
\frac{\partial \log P(\bfv; \alpha)}{\partial \alpha} = -\frac{\partial\mathcal{F}(\bfv; \alpha)}{\partial \alpha} - \frac{1}{\zeta(\alpha)}\frac{\partial \zeta(\alpha)}{\partial \alpha},
\end{equation}
where $\mathcal{F}(\bfv; \alpha) = - \log\left( \int e^{-E(\bfv, \bfh, \theta) - \phi(\theta; \alpha)} d\bfh d\theta \right)$ is called \textit{free energy}. It is easy to show that the gradients of the free energy and the partition function have the following form:
\begin{equation}
\label{eq:grad_free_energy}
\begin{aligned}
\frac{\partial\mathcal{F}(\bfv; \alpha)}{\partial \alpha} &=
\int P(\theta \mid \bfv; \alpha) \left( \frac{\partial \phi(\theta; \alpha)}{\partial \alpha} \right) d\theta,
&
\frac{1}{\zeta(\alpha)}\frac{\partial \zeta(\alpha)}{\partial \alpha} &=
-\int P(\theta; \alpha) \left( \frac{\partial \phi(\theta; \alpha)}{\partial \alpha} \right) d\theta.
\end{aligned}
\end{equation}
The final expression for the gradient has a contrastive divergence like form \citep{hinton2002training}:
\begin{equation}
\label{eq:grad_ll_EBSE_CD}
\frac{\partial \log P(\bfv; \alpha)}{\partial \alpha} = \mathbb{E}\left[\frac{\partial \phi(\theta; \alpha)}{\partial \alpha} \right]_{P(\theta; \alpha)} -
\mathbb{E}\left[\frac{\partial \phi(\theta; \alpha)}{\partial \alpha} \right]_{P(\theta \mid \bfv; \alpha)},
\end{equation}
where the expectations $\mathbb{E}[\ \cdot\ ]_{P(\theta; \alpha)}$ and $\mathbb{E}[\ \cdot\ ]_{P(\theta \mid \bfv; \alpha)}$ are taken over the marginal and conditional distributions over the models, respectively.
Based on the properties of the $P(\theta; \alpha)$ distribution, these expectations can be either fully estimated by Monte Carlo sampling, or partly computed analytically.

Since the expectations depend on the parameter $\alpha$, EBSE training is reminiscent of expectation-maximization (EM):
After initializing $\alpha$, in the E-step of the algorithm, we estimate $\mathbb{E}[\ \cdot\ ]_{P(\theta; \alpha)}$ and $\mathbb{E}[\ \cdot\ ]_{P(\theta \mid \bfv; \alpha)}$ using the current value of $\alpha$.
In the M-step, we maximize the log-likelihood by using gradient-based optimization.

\begin{algorithm}[!t]
\caption{Expectation-maximization k-step contrastive divergence for RBSE}\label{alg:RBSE-CD-k}
\begin{algorithmic}[1]
\Input $\bfS$---training (mini-)batch; learning rate $\lambda$; se-RBM(\bfv, \bfh, \bfW, \bfb, \bfc)
\Output gradient approximation $\Delta\alpha$ [depends on the parametrization of $\phi(\bfW, \bfb, \bfc; \alpha)$]
\State initialize $\Delta\alpha = 0$
\ForAll {the $\bfv \in \bfS$}
  \State $\bfv^{(0)} \leftarrow \bfv$
  \State $\bfh^{(0)} \leftarrow$ persistent state (or sample)
  \State sample $\bfW^{(0)}, \bfb^{(0)}, \bfc^{(0)} \sim P(\bfW, \bfb, \bfc \mid \bfv^{(0)}, \bfh^{(0)})$
  \For {$t = 0, \dots, k$} \Comment{CD-k for sampling from $P(\bfv, \bfh)$}
    \State sample $\bfW^{(t)}$, $\bfb^{(t)}$, $\bfc^{(t)} \sim P(\bfW, \bfb, \bfc \mid \bfv^{(t)}, \bfh^{(t)})$
    \State sample $\bfv^{(t)} \sim P(\bfv \mid \bfh^{(t)}, \bfW^{(t)}, \bfb^{(t)}, \bfc^{(t)})$
    \State sample $\bfh^{(t)} \sim P(\bfh \mid \bfv^{(t)}, \bfW^{(t)}, \bfb^{(t)}, \bfc^{(t)})$
  \EndFor
  \State $\Delta\alpha_m = \mathbb{E}\left[ \frac{\partial \phi}{\partial \alpha} \mid \bfv^{(k)}, \bfh^{(k)} \right]$
  \For {$t = 0, \dots, k$} \Comment{CD-k for sampling from $P(\bfh \mid \bfv)$}
    \State sample $\bfW^{(t)}$, $\bfb^{(t)}$, $\bfc^{(t)} \sim P(\bfW, \bfb, \bfc \mid \bfv, \bfh^{(t)})$
    \State sample $\bfh'^{(t)} \sim P(\bfh' \mid \bfv, \bfW^{(t)}, \bfb^{(t)}, \bfc^{(t)})$
  \EndFor
  \State $\Delta\alpha_c = \mathbb{E}\left[ \frac{\partial \phi}{\partial \alpha} \mid \bfv, \bfh'^{(k)} \right]$
  \State $\Delta\alpha \leftarrow \Delta\alpha + \lambda \left( \Delta\alpha_m - \Delta\alpha_c \right)$ \Comment{Estimate SGD step}
\EndFor
\State $\Delta\alpha = \Delta\alpha / \cord(\bfS)$
\end{algorithmic}
\end{algorithm}
\vspace{-1ex}

\subsection{Model Structure}\label{sec:model_strucutre}
We further consider a specific energy-based stochastic ensemble composed of restricted Boltzmann machines (RBM).
The energy function of the RBM is linear in each of the variables $\bfv$, $\bfh$, and $\theta$:
\begin{equation}
\label{eq:RBM_energy}
E(\bfv, \bfh, \theta) = - (\bfv^T \bfW \bfh + \bfb^T \bfv + \bfc^T \bfh)
= -(\sum_{i,j} W_{ij} v_i h_j + \sum_{i} b_i v_i + \sum_{j} c_j h_j),
\end{equation}
where parameters $\theta$ are represented by a tuple $(\bfW, \bfb, \bfc)$, and $\bfW \in \mathbb{R}^{D \times K}$, $\bfb \in \mathbb{R}^{D}$, $\bfc \in \mathbb{R}^{K}$.

RBM is an undirected graphical model that can be represented by two layers of interconnected probabilistic units (Figure~\ref{fig:RBM}).
These units can be seen as neurons with a probabilistic sigmoid activation function, and the graphical model can be represented by a two-layer neural network.
In this case, the parameters $\bfW$ play the role of connection strengths between the neurons.
By imposing a noise distribution on the parameters, connection strengths become stochastic.
From a graphical model perspective, this is equivalent to introducing new latent variables (Figure~\ref{fig:RBSE}).
Notice that the model becomes mixed directed and undirected---an ensemble of undirected RBMs generated by a simple Bayesian network composed of $(\bfW, \bfb, \bfc)$ random hidden variables.

In order to fully define the model, we need to choose a specific form of the marginal distribution $P(\theta; \alpha)$ that generates the ensemble.
First, to make the expectations computationally feasible, we suppose that $\bfW, \bfb, \bfc$ are all marginally independent:
\begin{equation}
\label{eq:RBSE_prob_factorisation}
P(\theta; \alpha)= \prod_{ij} P(W_{ij}; \alpha_{W_{ij}}) \prod_{i} P(b_i; \alpha_{b_i}) \prod_{j} P(c_j; \alpha_{c_j}).
\end{equation}
Then, we consider two cases:
(a) Bernoulli distributed parameters.
In this case $W_{ij}$ can be either zero with probability $1 - p_{ij}$, or equal to some $\bar W_{ij}$ with probability $p_{ij}$.
The tunable parameters are $\alpha_{ij} = \{p_{ij}, \bar W_{ij}\}$.
This case is similar to DropConnect \citep{wan2013regularization} technique but with adaptive distributions over the connections between visible and hidden layers in RBM.
(b) In the second case, parameters are normally distributed, i.e., $W_{ij} \sim \mathcal{N}(\mu_{ij}, \sigma_{ij})$, and $\alpha_{W_{ij}} = \{\mu_{ij}, \sigma_{ij}\}$.
In both cases, $\bfb$ and $\bfc$ are parametrized similarly as $\bfW$.
The number of parameters in the RBSE is twice the original number for the RBM.

This structure of the model (Figure~\ref{fig:RBSE}) implies the following set of independencies:
\begin{equation}
\label{eq:RBSE_cond_indep}
\begin{aligned}
v_i &\indep v_j \mid \bfh, & h_i &\indep h_j \mid \bfv, & W_{ij} &\indep W_{lk}, & W_{ij} &\nindep W_{ik} \mid v_i, & W_{ij} &\nindep W_{lj} \mid h_i.
\end{aligned}
\end{equation}
These expressions show the marginal independencies we purposefully incorporated into $P(\theta; \alpha)$ and conditional dependences between the components of $\bfW$.
Due to these dependencies between stochastic connections though visible and hidden units, the model is able to capture the data variance in the distribution over the stochastic parameters $\theta$.
Importantly, even though the components of $\bfW$ are dependent on each other given $\bfv$ and $\bfh$, we are still able to factorize the conditional distribution $P(\theta \mid \bfv, \bfh)$ using the renormalization procedure (see the supplementary material for details).

\subsection{Training}\label{sec:training}
We propose the expectation-maximization k-step contrastive divergence algorithm for training RBSE (summarized in Algorithm~\ref{alg:RBSE-CD-k}) with two different types of stochastic connections---Bernoulli and Gaussian---between the visible and the hidden layers.
To carry out the E-step, we need to compute the expectations in \eqref{eq:grad_ll_EBSE_CD}. We use Monte Carlo estimates of these expectations:
\begin{equation}
\label{eq:expectations_EBSE}
\begin{aligned}
&\mathbb{E}[\ \cdot\ ]_{P(\theta; \alpha)}
= \int d\bfv d\bfh~P(\bfv, \bfh; \alpha) \int d\theta~P(\theta \mid \bfv, \bfh; \alpha) [\ \cdot\ ]
\approx \frac{1}{M} \sum_{\tilde\bfv, \tilde\bfh \sim P(\bfv, \bfh; \alpha)} \int d\theta~P(\theta \mid \tilde\bfv, \tilde\bfh; \alpha) [\ \cdot\ ],\\
&\mathbb{E}[\ \cdot\ ]_{P(\theta \mid \bfv; \alpha)}
\approx \frac{1}{M} \sum_{\hat\bfh \sim P(\bfh \mid \bfv; \alpha)} \int d\theta~P(\theta \mid \bfv, \hat\bfh; \alpha) [\ \cdot\ ],
\end{aligned}
\end{equation}
where $M$ is the number of samples used to approximate a distribution. The states $(\tilde\bfv, \tilde\bfh)$ should be sampled from the marginal model distribution $P(\bfv, \bfh; \alpha)$, and $\hat\bfh$ sampled from the marginal model conditional distribution $P(\bfh \mid \bfv; \alpha)$.
This can be achieved by running a Gibbs sampling procedure as in standard contrastive divergence algorithm (see Algorithm~\ref{alg:RBSE-CD-k}).

Lastly, we need to compute the expectation over the posterior $P(\theta \mid \bfv, \bfh; \alpha)$ given a visible and a hidden state.
In both Bernoulli and Gaussian cases, due to the structure of the ensemble distribution $P(\theta; \alpha)$ discussed in section~\ref{sec:model_strucutre}, these expectations can be computed analytically.
As an example, we get the following expressions for the gradient update of the Gaussian RBSE (for the Bernoulli case, the notation and other details see the supplementary material):
\begin{equation}
\nonumber
\begin{aligned}
\frac{\partial \log P(\bfv; \alpha)}{\partial \bar W_{ij}} =
& \left< v_i h_i \right>_\text{data} - \left< v_i h_i \right>_\text{recon}, &
\frac{\partial \log P(\bfv; \alpha)}{\partial \bar \sigma_{ij}} =
& \left< v_i^2 h_j^2 \sigma_{ij} \right>_\text{data} - \left< v_i^2 h_j^2 \sigma_{ij} \right>_\text{recon}.
\end{aligned}
\end{equation}
For the models where the expectations over the ensemble are not analytically tractable, we can estimate them using Monte Carlo Markov chain sampling as well.

The main bottleneck of the algorithm is the number of samplings per gradient update.
Since all the connections between the visible and the hidden units are probabilistic, the random variables need to be sampled a quadratic number of times compared to learning an RBM.
However, due to the independencies introduced in Eq. (\ref{eq:RBSE_cond_indep}), all the variables $W_{ij}$, $b_i$, and $c_j$ can be sampled in parallel.

\section{Experiments}\label{sec:experiments}
We introduce the concept of stochastic representations by considering a semi-supervised learning scenario where a large amount of data is available but with only a few labeled examples per class.
RBSE is a generative model, and hence it can be trained in an unsupervised fashion on the entire dataset.
Once the ensemble distribution $P(\theta; \alpha)$ is tuned to the data, for every labeled training example, we can conditionally sample models $\theta \sim P(\theta \mid \bfv)$ and then use each model to generate a representation based on the activation of the hidden units $\bfh$.
In other words, the generative stochastic ensembles can be used to store the information about the entire dataset and effectively augment the number of labeled examples by generating multiple stochastic representations.
This is analogous to the concept of corrupted features \citep{maaten2013learning,wager2013dropout}.
The main difference is that RBSEs can learn from the unlabeled part of the dataset how to corrupt the data properly.

To test the concept, we implemented Bernoulli and a Gaussian RBSE using the Theano library \citep{bergstra2010theano}. We trained our generative models on two-dimensional synthetic datasets and on MNIST digits. The following sections provide the details on each of the experiments.

\begin{figure}[t!]
\centering
\begin{subfigure}[b]{0.49\textwidth}
\includegraphics[width=\textwidth]{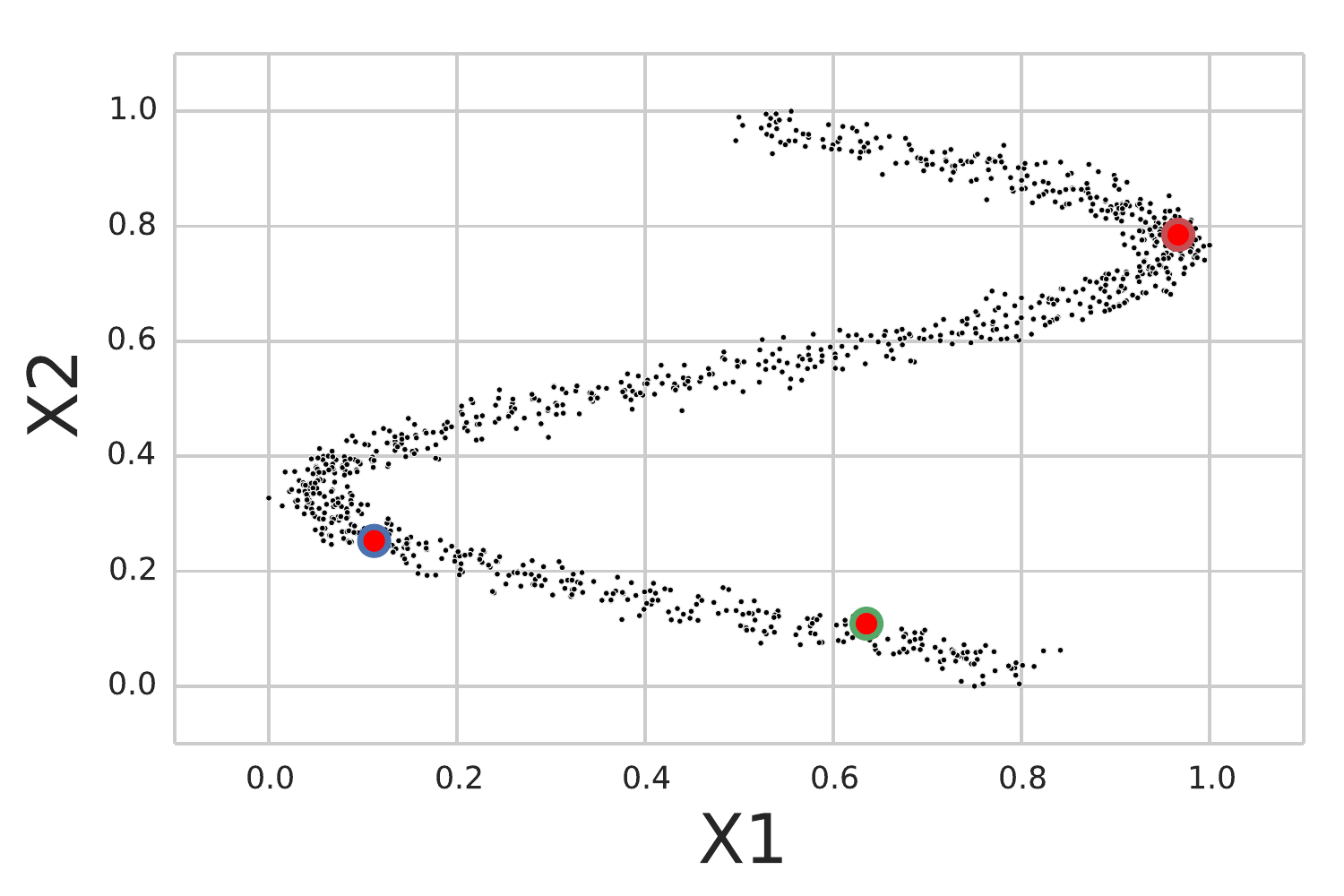}
\caption{RBM performs deterministic mapping.}
\label{fig:RBM_synthetic}
\end{subfigure}
\begin{subfigure}[b]{0.49\textwidth}
\includegraphics[width=\textwidth]{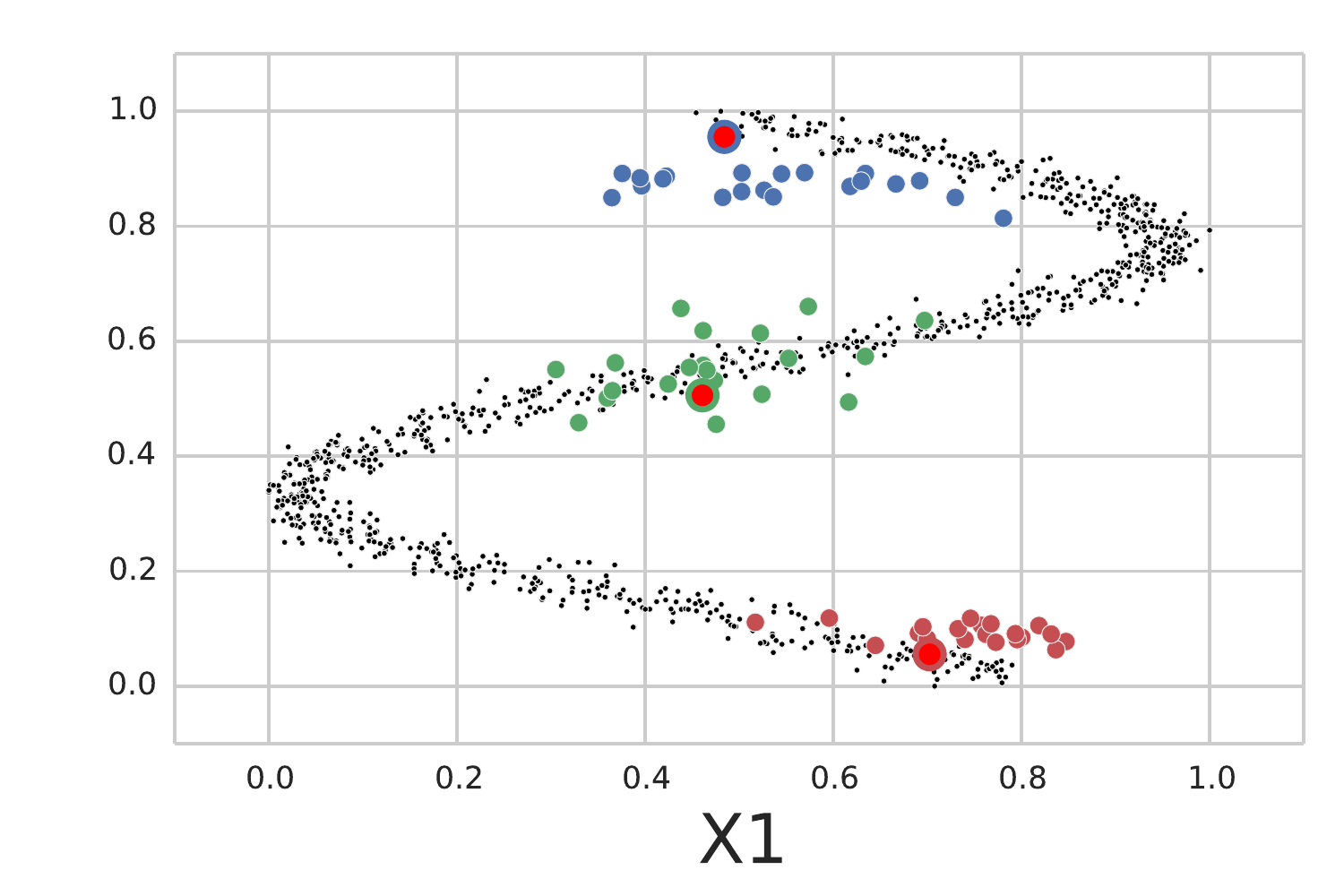}
\caption{RBSE captures the training data variance.}
\label{fig:RBSE_synthetic_a}
\end{subfigure}
\begin{subfigure}[b]{0.49\textwidth}
\includegraphics[width=\textwidth]{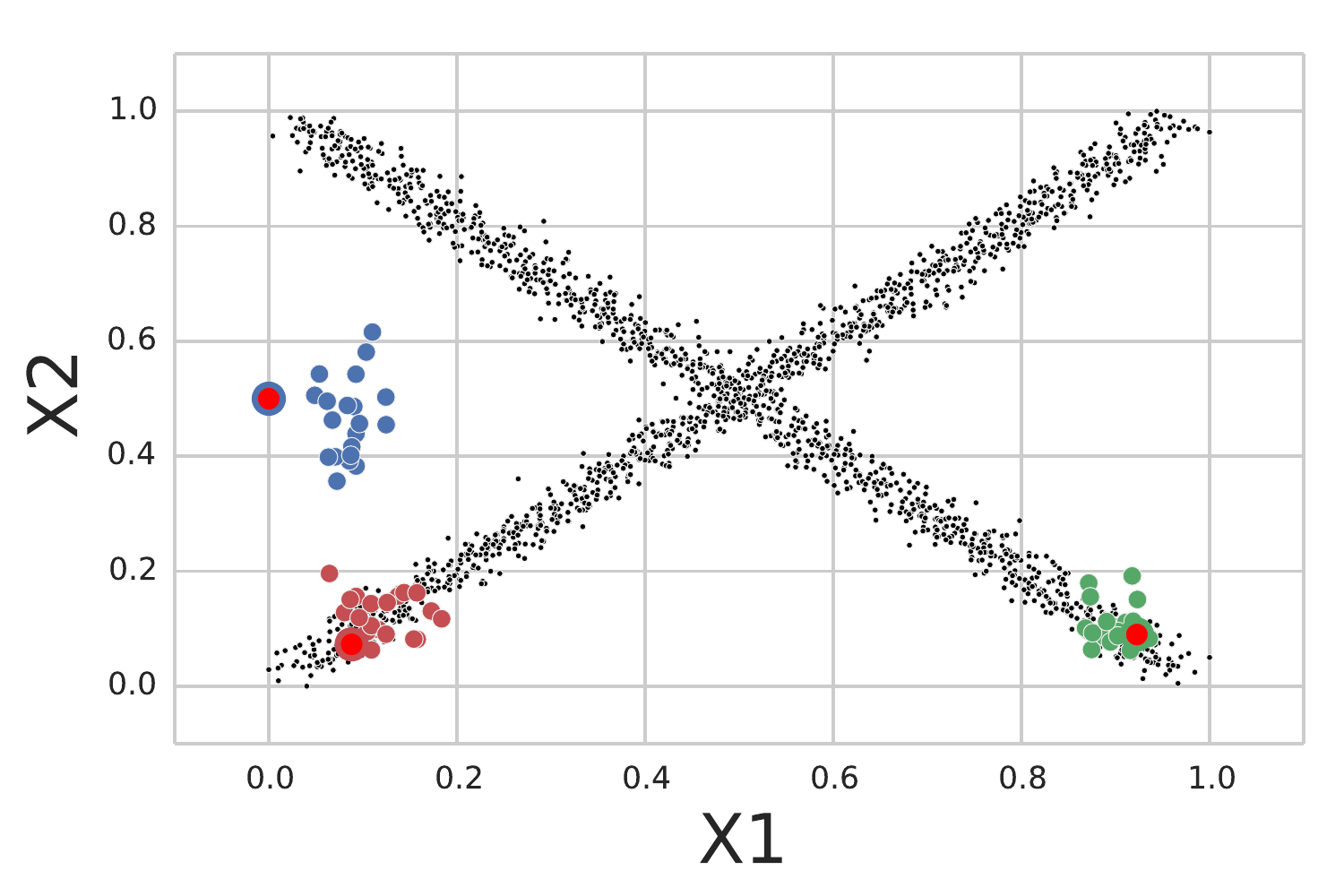}
\caption{Isolated points are attracted by the training data.}
\label{fig:RBSE_synthetic_b}
\end{subfigure}
\begin{subfigure}[b]{0.49\textwidth}
\includegraphics[width=\textwidth]{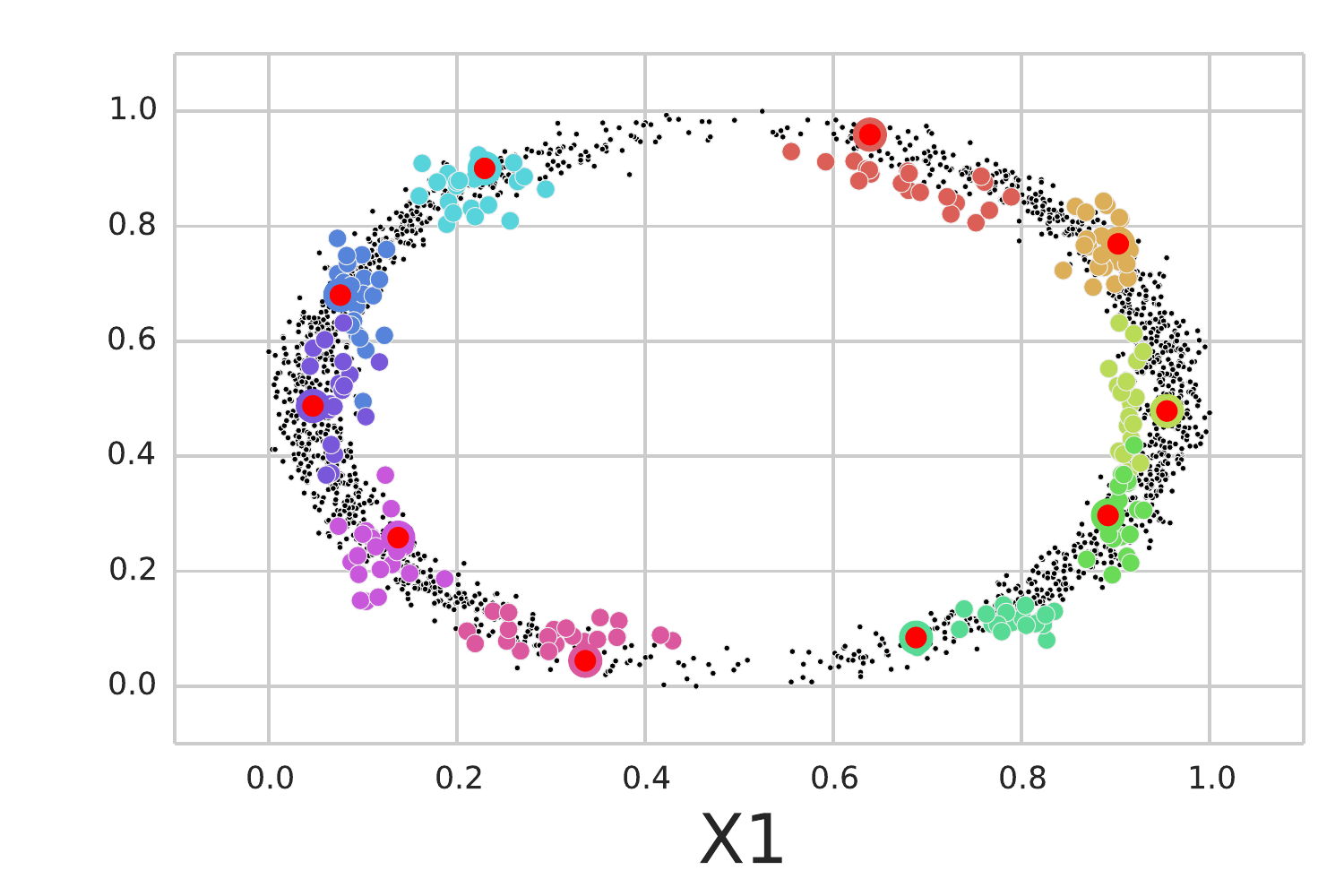}
\caption{RBSE captures the whole data manifold.}
\label{fig:RBSE_synthetic_c}
\end{subfigure}
\caption{Visualization of the experiments with the synthetic two dimensional data.
On all panels, the black points represent the training data, and the red points are the testing. Best viewed in color.
(a) The classical RBM can perform only deterministic mapping.
(b) RBSE maps the testing points to multiple representations which map backwards to different points in the original space.
(c) RBSE attracts the representations of the isolated points towards the training data manifold.
(d) Stochastic representations can capture the variance of the entire manifold from a few examples.
\vspace{-2ex}
}
\label{fig:syntheic_experiments}
\end{figure}

\subsection{Synthetic Data}\label{subsec:synthetic_data}
In order to visually test how a stochastic ensemble can capture the training data manifold, we generated several simple two-dimensional datasets $\{x_1, x_2\} \in [0, 1]^2$ (Figure~\ref{fig:syntheic_experiments}).
We trained an ordinary RBM with 2 visible and 8 hidden units and an RBSE of the same structure.
Using these models, we mapped the testing points to the latent space multiple times, and then back to the original space.

For RBM, we used the mean field activations for new representations: $\bfh_{new} = P(\bfh \mid \bfv)$ and $\bfv_{new} = P(\bfv \mid \bfh_{new})$.
Unsurprisingly, the two consecutive transformations, from visible to hidden and back to visible space, performed by a properly trained RBM always mapped the testing points to themselves (Figure~\ref{fig:RBM_synthetic}).
Notice that this holds not only for RBMs but for any deterministic model:
Other point-wise deterministic representation learning techniques, e.g., autoencoders \citep{bengio2009learning}, exhibit the same behavior.

We performed a similar procedure multiple times for RBSE: first, we sampled a model from the conditional distribution $P(\theta \mid \bfv)$, then using $P(\bfh \mid \bfv, \tilde\theta)$, we transformed testing points from the visible to the hidden space, then mapped backwards using the average model from the ensemble.
Stochastic representations of the testing data, when mapped back to the visible space, were distributed along the training data manifold near the testing points they belonged to (Figure~\ref{fig:RBSE_synthetic_a}).

The experiments also demonstrated that the representations of an isolated testing point (an outlier) will be attracted towards the manifold captured by RBSE (Figure~\ref{fig:RBSE_synthetic_b}), which cannot be done by the RBM.
Finally, an entire manifold can be captured by the generated stochastic representations for only a small number of initial data points (Figure~\ref{fig:RBSE_synthetic_c}).
These experiments confirmed the capability of RBSEs to refine its internal distribution over the models and to capture the variance of the data.
Therefore, the stochasticity of such tunable ensembles should provide better regularization than just arbitrary noise injection and, as we show further, improve the perfomrance in one-shot learning.

\begin{figure}[t!]
\begin{subfigure}[b]{0.28\textwidth}
\includegraphics[width=\textwidth]{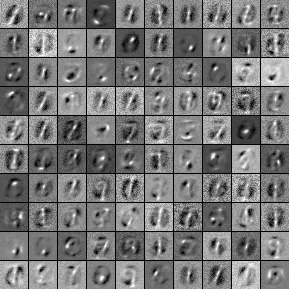}
\caption{RBSE filters}
\label{fig:RBSE_filt}
\end{subfigure}
~
\begin{subfigure}[b]{0.28\textwidth}
\includegraphics[width=\textwidth]{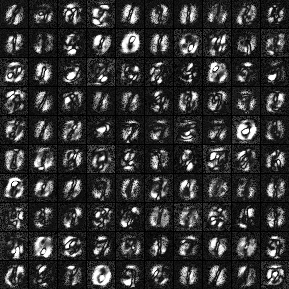}
\caption{RBSE probabilities}
\label{fig:RBSE_prob}
\end{subfigure}
~
\begin{subfigure}[b]{0.43\textwidth}
\includegraphics[width=\textwidth]{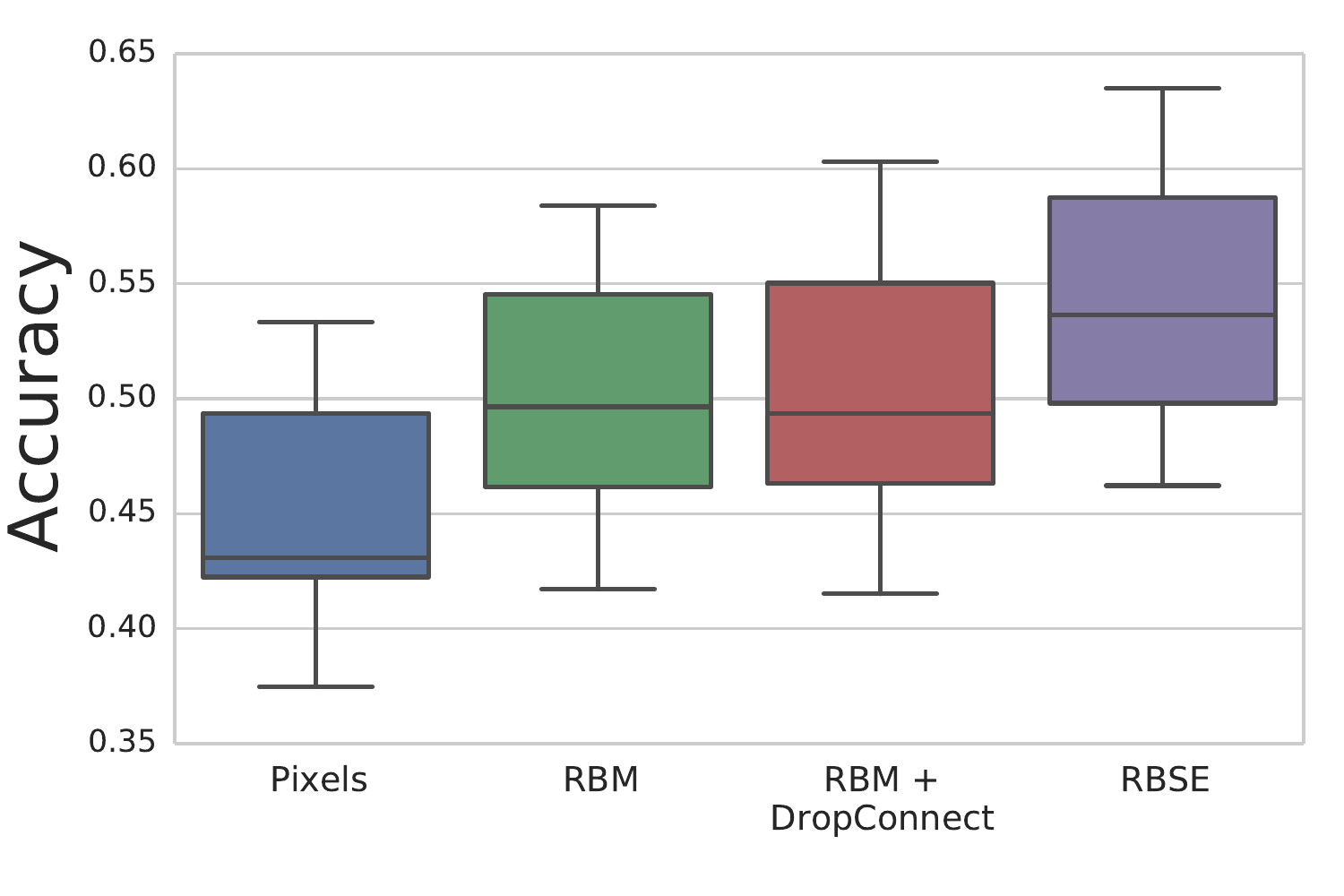}
\caption{One-shot accuracy on 10-class classification}
\label{fig:one-shot}
\end{subfigure}
\caption{After training on MNIST for several epochs: (a) RBSE filters, i.e., the $\bar W_{ij}$ values. (b) RBSE connection probabilities, i.e., the $p_{ij}$ values---the dark pixels are close to 0; the light pixels are close to 1. (c) Performance of a logistic regression classifier trained on different representations of the data under the one-shot constraints.}
% \vspace{-2ex}
\label{fig:MNIST_results}
\end{figure}

\subsection{One-Shot Learning of MNIST Digits}\label{subsec:mnist}
To test stochastic representations in the semi-supervised setting where only a few labeled data are available (one example per class), we performed experiments on MNIST digits dataset.
We trained a Bernoulli RBSE model with 784 visible and 400 hidden units on 50,000 unlabeled MNIST digits using the proposed Algorithm~\ref{alg:RBSE-CD-k} with a number of MCMC steps $k=1$.
The learned filters (i.e., $\bar W_{ij}$) and the Bernoulli connection probabilities (i.e., $p_{ij}$) arranged into tiles are presented on Figure~\ref{fig:MNIST_results}.
Notice that the connection probabilities encode a lot of structure (Figure~\ref{fig:RBSE_prob}).
For the purpose of comparison, we also trained an RBM of the same configuration on the same unlabeled MNIST data.

Next, we sampled 1000 objects (100 per class) from the remaining 20,000 MNIST digits: 1 training and 99 testing examples per class.
For every training sample we used different approaches to construct the representations: (a) image pixels, (b) deterministic representations constructed by a pre-trained RBM, (c) stochastic representations constructed by an RBM with Bernoulli probabilities 0.5 for every connection (equivalent to DropConnect), and (d) RBSE-generated stochastic representations.
In the (c) and (d) cases, we constructed 10 representations for every object.

Finally, we trained and tested a logistic regression classifier (with no additional regularization) on the these representations under one-shot learning constraints.
The classification experiments were done multiple times for differently sampled objects.
The results are presented on Figure~\ref{fig:one-shot}.
About 10\% improvement in classification accuracy is due to better representations learned by RBM (disentanglement of the classes).
When we regularize the classifier by generating stochastic representations with DropConnect noise applied to trained RBM, the performance slightly drops.
On the contrary, when the classifier is regularized through the representations generated by RBSE, we get about another 5\% increase in accuracy on average.

\section{Discussion}\label{sec:discussion}
In this paper, we introduced the concept non-deterministic representations that can be learned by Energy-based Stochastic Ensembles tuned to the data in unsupervised fashion.
The ensemble stochasticity can capture the data variance and be used to adaptively regularize discriminative models and improve their performance in semi-supervised settings.
The actual learning of a proper model perturbation from the data is the conceptual novelty compared to previous work \citep{bachman2014learning,maaten2013learning,wager2013dropout}.
We illustrated the power of stochastic ensembles visually on synthetic two-dimensional data and demonstrated it quantitatively on one-shot learning of the MNIST hand-written digits.

The inspiration from synaptic stochasticity observed in biological nerve cells provides a number of insights and hypotheses for experimental neuroscience, which we will report separately elsewhere.
From the artificial neural networks perspective, the proposed approach of using stochastic connections between neural units is interesting as well.
For example, similar stochastic ensembles of feedforward neural networks should be able to capture complex multi-modal data manifolds \citep{tang2013learning}.
Also, recently proposed Generative Stochastic Networks (GSNs), which are used to encode probabilities in their sampling behavior \citep{bengio2013gsn,bengio2013generalized}, can be naturally endorsed with non-deterministic connections and might potentially realize richer families of distributions.

Interestingly, biological inspiration also suggests that neuromorphic computers that operate in a massively parallel fashion, while consuming a faction of the power of digital computers \citep{Merolla_etal14} can be leveraged.
They often natively support Bernoulli synaptic stochasticity~\citep{Goldberg_etal01}, and neuromorphic variants of RBMs can be efficiently implemented and trained \citep{Neftci_etal14}.
This suggests that the disadvantages associated to the computational overhead of RBSEs can be nullified by using an appropriate computational substrate.

\subsubsection*{Acknowledgments}
The authors thank Cian O'Donnell for helpful discussions.
E.N. and G.C. were supported in part by the National Science Foundation (NSF EFRI-1137279) and the Office of Naval Research (ONR MURI 14-13-1-0205).
M.A. was supported by KAUST Graduate Fellowship.

\bibliography{references}
\bibliographystyle{iclr2015}

\end{document}

\maketitle

\appendix
% \section*{APPENDIX}

\section{Posterior Bernoulli and Gaussian Distributions}
\renewcommand{\theequation}{A\arabic{equation}}
\setcounter{equation}{0}

This section provides details on the posterior distributions for Bernoulli and Gaussian stochastic ensembles.
RBM has three sets of parameters: connection strengths between the visible and hidden layers $\bfW \in \mathbb{R}^{D \times K}$, bias $\bfb \in \mathbb{R}^{D}$ for visible, and bias $\bfc \in \mathbb{R}^{K}$ for hidden units.
We use $i$ to index $D$ visible dimensions, and $j$ for $K$ hidden dimensions.
We denote unnormalized measures by $\tilde P(\cdot)$.
When the equations for $W_{ij}$, $b_i$, and $c_j$ have the same form, we refer to these components as $\theta_k$.

For some fixed $(\bfv, \bfh)$ we know that RBSE's energy is a linear function of $\theta$.
Since the prior $P(\theta; \alpha)$ factorizes over all the components of $\theta$ according to our definition.
Thus, we can factorize unnormalized $\tilde P(\theta, \bfv, \bfh; \alpha)$ over all $\theta_k$:
\begin{equation}
\label{eq:bernoulli_unnormalized}
\begin{aligned}
\tilde P(W_{ij}, v_i, h_j; \alpha_{ij}) &= P(W_{ij}; \alpha_{ij}) \exp(v_i h_j W_{ij}),\\
\tilde P(b_i, v_i; \alpha_i) &= P(b_i; \alpha_i) \exp(v_i b_i),\\
\tilde P(c_j, h_j; \alpha_j) &= P(c_j; \alpha_j) \exp(h_j c_j).
\end{aligned}
\end{equation}
We can further renormalize \eqref{eq:bernoulli_unnormalized} and obtain the following posterior distribution:
\begin{equation}
\label{eq:bernoulli_posterior}
\begin{aligned}
P(\bfW, \bfb, \bfc \mid \bfv, \bfh; \alpha)
&= \frac{\prod_{ij} P(W_{ij}, v_i, h_j; \alpha_{ij}) \prod_i P(b_i, v_i; \alpha_i) \prod_j P(c_j, h_j; \alpha_j)}{\int \prod_{ij} P(W_{ij}, v_i, h_j; \alpha_{ij}) d\bfW \prod_i P(b_i, v_i; \alpha_i) d\bfb \prod_j P(c_j, h_j; \alpha_j) d\bfc}\\
&= \prod_{ij}\frac{P(W_{ij}, v_i, h_j; \alpha_{ij})}{\int P(W_{ij}, v_i, h_j; \alpha_{ij}) dW_{ij}} \prod_i\frac{P(b_i, v_i; \alpha_i)}{\int P(b_i, v_i; \alpha_i) db_i} \prod_j\frac{P(c_j, h_j; \alpha_j)}{\int P(c_j, h_j; \alpha_j) dc_j}\\
&= \prod_{ij} P(W_{ij} \mid v_i, h_j; \alpha_{ij}) \prod_{i} P(b_i \mid v_i; \alpha_{i}) \prod_{j} P(c_j \mid h_j; \alpha_{j}).
\end{aligned}
\end{equation}
As we see, if the prior distribution $P(\theta; \alpha)$ factorizes over the components of $\theta$, the posterior distribution $P(\theta \mid \bfv, \bfh; \alpha)$ is also factorisable.
Finally, we need to find the posterior distribution for each of the components for the Bernoulli and Gaussian ensembles.

For Bernoulli case, a priori, every component $\theta_k$ can either take some non-zero value $\bar\theta_k$ with probability $(1 - p_k)$, or be equal to zero with probability $p_k$.
Then, the distribution $P(\theta; \alpha)$ is parametrized by $\alpha = (\bar\theta_k, p_k)$ and has the following form:
\begin{equation}
\label{eq:bernoulli_prior}
P(\theta; \alpha) = \prod_k P(\theta_k; \alpha_k) = \prod_k \left[ \delta(\theta_k)(1 - p_k) + \delta(\theta_k - \bar\theta_k)p_k \right],
\end{equation}
where $\delta(\cdot)$ is the Dirac delta function. The posterior for $W_{ij}$ (similar for $b_i$ and $c_j$) will be
\begin{equation}
\label{eq:bernoulli_Wij_posterior}
P(W_{ij} \mid v_i, h_j; \alpha_{ij}) =
\frac{\left( \delta(W_{ij})(1 - p_{ij}) + \delta(W_{ij} - \bar W_{ij})p_{ij} \right) \exp(v_i h_j W_{ij})}{1 - p_{ij} + p_{ij}\exp(v_i h_j W_{ij}) }
\end{equation}

For Gaussian case, components $\theta_k$ a priori have a Gaussian distribution with the mean $\bar\theta_k$ and variance $\sigma_k^2$, i.e., $\alpha_k = (\bar\theta_k, \sigma_k)$.
Omitting the details of integration, the posterior for $W_{ij}$ will have the following form (similar expressions are easy to derive for $b_i$ and $c_j$):
\begin{equation}
\label{eq:gaussian_Wij_posterior}
\begin{aligned}
P(W_{ij} \mid v_i, h_i; \alpha_{ij})
= \frac{1}{\sqrt{2\pi}\sigma_{ij}} \exp\left( - \left[ \frac{W_{ij} - (\bar W_{ij} + v_i h_i \sigma_{ij}^2)}{\sqrt{2}\sigma_{ij}} \right]^2 \right).
\end{aligned}
\end{equation}

In both Bernoulli and Gaussian cases, due to the linear structure of the RBSE energy function and the structure of the prior $P(\theta; \alpha)$, we are able to get analytical expressions for the posterior of each component of $\theta$.
Moreover, the derivations will be the same not only for Bernoulli and Gaussian cases, but for any $P(\theta; \alpha)$ that is completely factorisable over $\{\theta_k\}$.

\section{Expectations Over The Posterior}
\renewcommand{\theequation}{B\arabic{equation}}
\setcounter{equation}{0}

Here we provide details on the analytical computation of the expected stochastic gradient update for the two types of stochastic ensembles: Bernoulli and Gaussian.
For the sake of space, as in Appendix~A, when there is no ambiguity, we denote the components of $\bfW, \bfb, \bfc$ generally as $\theta_k$.

The general form of the gradient expectation is the following (see Eq. 10 in the main text):
\begin{equation}
\label{eq:posterior_expected_grad}
\int_\Theta P(\theta \mid \bfv, \bfh; \alpha) \frac{\partial \phi(\theta; \alpha)}{\partial \alpha} d\theta.
\end{equation}

Now, we start with the Bernoulli case, where $\alpha_k = (\theta_k, p_k)$.
The prior distribution and the $\phi$ potential have the following form:
\begin{equation}
\label{eq:bernoulli_prior_and_phi_deriv}
\begin{aligned}
& P(\theta_k; \alpha_k) \equiv e^{-\phi(\theta_k; \alpha_k)} = \delta(\theta_k)(1 - p_k) + \delta(\theta_k - \bar\theta_k)p_k,\\
& \frac{\partial \phi}{\partial p_k} = \frac{\delta(\theta_k) - \delta(\theta_k - \bar\theta_k)}{e^{-\phi(\theta_k; \alpha_k)}},
\quad
\frac{\partial \phi}{\partial \bar\theta_k} = \frac{\delta'(\theta_k - \bar\theta_k)p_k}{e^{-\phi(\theta_k; \alpha_k)}},
\end{aligned}
\end{equation}
where $\delta(\cdot)$ is the Dirac delta function, and $\delta'(\cdot)$ is its derivative. When we plug \eqref{eq:bernoulli_posterior} and \eqref{eq:bernoulli_prior_and_phi_deriv} into \eqref{eq:posterior_expected_grad}, we obtain the following expressions for the expectation over $W_{ij}$ (similar for $b_i$ and $c_i$):
\begin{equation}
\label{eq:bernoulli_expected_grad_Wij}
\begin{aligned}
\mathbb{E}\left[ \frac{\partial \phi}{\partial p_{ij}} \mid v_i, h_i \right]
&= \frac{1 - p_{ij}}{1 - p_{ij} + p_{ij}\exp(v_i h_i \bar W_{ij})}\left(1 - e^{v_i h_i \bar W_{ij}}\right),\\
\mathbb{E}\left[ \frac{\partial \phi}{\partial \bar W_{ij}} \mid v_i, h_i \right] &= \frac{p_{ij}}{1 - p_{ij} + p_{ij}e^{v_i h_i \bar W_{ij}}} \int \delta'(W_{ij} - \bar W_{ij})\exp(v_i h_i W_{ij}) dW_{ij}\\
&= [\text{integrating by parts}] = \frac{p_{ij} e^{v_i h_i \bar W_{ij}}}{1 - p_{ij} + p_{ij}\exp(v_i h_i \bar W_{ij})} (- v_i h_i).
\end{aligned}
\end{equation}
Eventually, we get the following expressions for the gradient components of the log-likelihood:
\begin{equation}
\label{eq:bernoulli_RBSE_ll_grad}
\begin{aligned}
\frac{\partial \log P(\bfv; \alpha)}{\partial \bar W_{ij}} =
& \left< P\left( W_{ij} \neq 0 \mid v_i, h_i \right) v_i h_i \right>_\text{data} - \left< P\left( W_{ij} \neq 0 \mid v_i, h_i \right) v_i h_i \right>_\text{recon},\\
\frac{\partial \log P(\bfv; \alpha)}{\partial \bar p_{ij}} =
& \left< P\left( W_{ij} = 0 \mid v_i, h_i \right) \left(1 - e^{v_i h_i \bar W_{ij}}\right) \right>_\text{data} -\\
& \left< P\left( W_{ij} = 0 \mid v_i, h_i \right) \left(1 - e^{v_i h_i \bar W_{ij}}\right) \right>_\text{recon}.
\end{aligned}
\end{equation}
The gradient over $\bar W_{ij}$ resembles the original contrastive divergence but has additional posterior probability multipliers.

Similarly, we do the same derivations for the Gaussian case, where $\alpha_k = (\bar\theta_k, \sigma_k)$. The prior and the $\phi$ derivatives have the following form:
\begin{equation}
\label{eq:Gaussian_prior_and_phi_deriv}
\begin{aligned}
& P(\theta_k; \alpha_k) \equiv e^{-\phi(\theta_k; \alpha_k)} = \frac{1}{\sqrt{2\pi}\sigma_k} \exp\left[ -\left( \frac{\theta_k - \bar\theta_k}{\sqrt{2}\sigma_k} \right)^2 \right],\\
& \frac{\partial \phi}{\partial \bar\theta_k} = \frac{\bar\theta_k - \theta_k}{\sigma_k^2}, \quad
\frac{\partial \phi}{\partial \sigma_k} = \frac{1}{\sigma_k}\left[ 1 - \left( \frac{\theta_k - \bar\theta_k}{\sigma_k} \right)^2 \right].
\end{aligned}
\end{equation}
After marginalization, we obtain the following expressions for $W_{ij}$ (similar for $b_i$ and $c_j$):
\begin{equation}
\label{eq:gaussian_expected_grad_Wij}
\begin{aligned}
&\mathbb{E}\left[ \frac{\partial \phi}{\partial \bar W_{ij}} \mid v_i, h_i \right] = -v_i h_i, & \quad &
\mathbb{E}\left[ \frac{\partial \phi}{\partial \sigma_{ij}} \mid v_i, h_i \right] = - v_i^2 h_i^2 \sigma_{ij},
\end{aligned}
\end{equation}
and get the following expressions for the gradient of the log-likelihood components:
\begin{equation}
\label{eq:gaussian_RBSE_ll_grad}
\begin{aligned}
\frac{\partial \log P(\bfv; \alpha)}{\partial \bar W_{ij}} =
& \left< v_i h_i \right>_\text{data} - \left< v_i h_i \right>_\text{recon},\\
\frac{\partial \log P(\bfv; \alpha)}{\partial \bar \sigma_{ij}} =
& \left< v_i^2 h_j^2 \sigma_{ij} \right>_\text{data} - \left< v_i^2 h_j^2 \sigma_{ij} \right>_\text{recon}.
\end{aligned}
\end{equation}

Notice that when $\bfW$, $\bfb$, $\bfc$ are deterministic (i.e., $p_{ij} = 0$ and $\sigma_k = 0$ for Bernoulli and Gaussian ensembles, respectively), the expected derivatives over $\bar W_{ij}$ (the same is for $\bar b_i$ and $\bar c_j$) give the same expressions as the standard contrastive divergence.
Thus, RBM is the deterministic limit of RBSE.

Another point is related to the implementation: During the optimization, the algorithm steps might be suboptimal, hence we restrict probabilities be in the range $[\epsilon, 1-\epsilon]$ for some $\epsilon \ll 1$.
In other words, we do not allow the model occasionally turn into classic RBM by keeping some $\epsilon$-stochasticity in the connection strengths.